# Classification of Alzheimer's Disease with Deep Learning on Eye-tracking Data[*]

Harshinee Sriram[†]
Department of Computer Science
The University of British Columbia
Vancouver, BC, Canada
hsriram@cs.ubc.ca

Cristina Conati
Department of Computer Science
The University of British Columbia
Vancouver, BC, Canada
conati@cs.ubc.ca

Thalia Field
Vancouver Stroke Program and Division of Neurology, The University of British Columbia,
Vancouver, BC, Canada
thalia.field@ubc.ca

**ABSTRACT**

Existing research has shown the potential of classifying Alzheimer's Disease (AD) from eye-tracking (ET) data with classifiers that rely on task-specific engineered features. In this paper, we investigate whether we can improve on existing results by using a Deep Learning classifier trained end-to-end on raw ET data. This classifier (VTNet) uses a GRU and a CNN in parallel to leverage both visual (V) and temporal (T) representations of ET data and was previously used to detect user confusion while processing visual displays. A main challenge in applying VTNet to our target AD classification task is that the available ET data sequences are much longer than those used in the previous confusion detection task, pushing the limits of what is manageable by LSTM-based models. We discuss how we address this challenge and show that VTNet outperforms the state-of-the-art approaches in AD classification, providing encouraging evidence on the generality of this model to make predictions from ET data.

## 1 Introduction

In recent years, eye-tracking has been extensively investigated as a source of information for AI agents to determine relevant properties of their users. This research has already generated very encouraging results, showing that eye-tracking (ET) data can be used to train classifiers for predicting user short-term states such as confusion, affect, and mind wandering [26,4,25], as well as long-term properties such as cognitive abilities, personality traits, and health conditions [20,22,41]. Most of these results have been achieved by using traditional Machine Learning (ML) classifiers rather than Deep Learning (DL) methods. This is in part because existing ET datasets are usually relatively small, as acquiring accurate ET data currently requires specialized equipment and collection in a lab setting.

However, there have been some initial attempts to use DL methods to make predictions on user properties from ET data. Most of this existing work converts the ET data into a visual representation (i.e., scanpath or heatmap) that is then analyzed by a CNN-based classifier for prediction [5,10,30]. In contrast, Pusiol et al. [36] trained an RNN model on sequences of ET fixations (namely clusters of raw ET samples associated with unique points of attention) to distinguish between two developmental disorders.

A DL architecture that leverages both the visual (V) and temporal (T) aspects of ET data, called VTNet, was proposed by Sims and Conati [39]. VTNet includes a GRU and a CNN that operate in parallel, with the GRU taking as direct input raw sequential ET samples, while the CNN processes the corresponding scanpath image, namely a representation of the samples' X and Y coordinates and the transitions between them. This approach was successfully used for classifying user confusion while processing visualizations [26]. The approach also outperformed its GRU and CNN components when they were trained, respectively, on the temporal and visual representation of the ET sequences, showing the value of combining the two representations.

In this paper, we investigate if the VTNet architecture can also improve on the state-of-the-art results in a very different context: leveraging ET data to classify Alzheimer's Disease (AD). There has been increasing interest in devising lightweight classifiers of AD as an initial screening for this condition [15,17,32] because existing assessments tend either to be resource-intensive and time-consuming (e.g., specialized neuroimaging and detailed cognitive assessments), or they are lightweight cognitive screening tools [7] that are not sensitive enough to detect AD or other mild cognitive impairments that can develop into AD. There is existing research that shows the potential of classifying AD from ET data alone or together with language data generated during simple screening tasks [3,20,35]. Most of these works use non-DL classifiers, and they all rely on features engineered based on knowledge of the task at hand.

In this paper, we investigate if VTNet can improve these results when trained end-to-end from raw ET data. In particular, we focus on the previous work that currently has the most reliable results [20] obtained from one of the largest ET datasets for AD screening in the literature (AD dataset from now on), which was collected from participants engaging in three different tasks: a pupil calibration task, a picture description task, and a textual paragraph reading task. The challenge in working with this dataset is that the ET data sequences are much longer than those leveraged in the previous work that used VTNet for classifying confusion. Sequence length is known to be a potentially limiting factor in the effectiveness of LSTM-based models if the length is beyond 1,000 timesteps [27], and the AD dataset involves sequences that have an average length of over 8,000 timesteps, with a maximum of over 26,000 timesteps. We address this issue by first exploring ways to reduce the sequence lengths, and next by augmenting the VTNet architecture with an attention layer, given that attention has been successfully used in Natural Language Processing (NLP) and Computer Vision to allow a model to focus on the most important parts of an input sequence.

Our results show that combining targeted length reduction with the addition of the attention layer allows VTNet to outperform state-of-the-art AD classifiers. These results entail two contributions. The first contribution is a step forward in the quest for accurate lightweight classifiers for AD. The second contribution is that we show the value of the VTNet architecture in a very different classification task, thus providing initial evidence on the generality of this approach for improving the ability of AI agents to leverage ET data to make classifications on relevant properties of their users.

The rest of this paper is organized as follows: Section 2 reviews related work. Sections 3 and 4 describe the AD dataset and the data preprocessing steps. Section 5 summarizes the VTNet model architecture. Section 6 evaluates VTNet with ET sequences of different lengths, while Section 7 evaluates VTNet with attention. Finally, Section 8 concludes the paper and discusses avenues for future work.

## 2 Related Work

**Leveraging ET data for AD classification**. Research on ET data as a source of information for AD detection has been inspired by evidence that AD affects the functioning of the eye, causing abnormalities in fixations, saccades, and pupillary responses [15,31,32]. All existing works trained classifiers on ET data features based on knowledge of what was important during the visual tasks designed to test functionalities known to be degraded by AD. For instance, Pavisic et al. [35] trained their classifiers on features relevant to assess one's performance during tasks that tested fixation stability, focus on appearing stimuli and tracking a moving target. Biondi et al. [3] leveraged ET features relevant to reading tasks (e.g. number of repeated fixations on a word). Jang et al. [20] leveraged ET data from tasks such as pupil calibration, picture description, and reading, and trained their classifiers on features defined by looking at specific regions of interest (ROIs) in their tasks (e.g., parts of the picture and paragraph).

All these works reported accuracies above 80%, however, the datasets used in [35] and [3] were rather small in size (57 and 69 datapoints respectively), and these works only reported classification accuracy as their performance metric. On the other hand, Jang et al. [20] leveraged a larger dataset (126 datapoints) and provided an extensive analysis based on several performance metrics. Therefore, we use the data and classifiers in Jang et al. [20] to test and compare the performance of our proposed VTNet approach, which, in contrast to previous approaches, is trained directly on the raw ET samples, hence removing the need to create task-specific engineered features.

**DL classifiers leveraging ET data.** Recently, several studies have employed DL techniques to make predictions about users from their ET data [1,5,10,30,36,43]. Most of these studies convert the ET data into a visual representation, namely a *scanpath* or a *heatmap* (a format that uses color to show the degree of attention to a visual display), which is then used by a CNN-based classifier for prediction. For instance, CNN models have used scanpaths to predict the strategies of participants playing games (e.g., chess [30] and different types of economic games [5]). CNN models trained on heatmaps have been used to classify the ages of toddlers viewing images [10] and the attentional states of participants performing tasks in Augmented Reality [43]. In all these works, the CNN classifiers outperformed a non-DL baseline. Because these approaches do not process the sequences of ET data, they do not face the problem of excessive sequence length.

In contrast to approaches that rely on the visual representation of ET data, Pusiol et al. [36] trained their classifiers on the ET sequences themselves. The data was collected from participants with two different developmental disorders, and this work proposed an RNN-based classifier to distinguish between the two disorders. The training data consisted of ET sequences indicating if a patient was looking at certain face regions (nose, jaw, etc.) of a practitioner who was conducting a diagnostic interview with them, where the sequences were obtained by overlaying the participant's fixation points on the video of the interviewer's face. The resulting sequences were 3,000 timesteps in length. The authors experimented with three window lengths (15, 50, and 250 timesteps), to determine how much sequential information is required for classification and found the RNN trained on the 250 timesteps sequences to be the winning classifier.

Finally, Asish et al. [1] leveraged sequences of raw ET data to classify the distraction level of students in a Virtual Reality classroom. The sequences in the dataset were between 12,000 and 33,000 timesteps long. The authors compared a CNN, an LSTM, and a sequential combination of the two, trained on the data end-to-end, against a Random Forest (RF) classifier trained on summary statistics of the data, with the RF classifier being the winning model.

**Leveraging attention to focus on important parts of long sequences.** In Natural Language Processing (NLP), several works have used attention to extract important parts from sentences, paragraphs, and utterances for tasks such as speech recognition [6], dialogue act detection and key term extraction [38], and relation extraction [48]. The average length of sequences in these datasets ranges from 36 words per sentence [48] to 120 words per article [38]. Later work in NLP relies on transformer-based architectures to deal with longer sequences (e.g., [9]), but transformers are unsuitable for our work because they are complex models that require large datasets for training.

Attention has also been used in Computer Vision to extract important parts from videos for tasks such as video summarization [21,37], skill level assessment [13], and video question answering [45]. The datasets used in these works originally range from 15,000 timesteps [21] to 540,000 timesteps

[37]. To deal with such high sequence lengths, all these approaches reduce the length by decreasing the video sampling frequency [13,21,37,45]. In addition, in [37], the authors use knowledge-based approaches to extract interesting video segments [37], whereas in [13] they uniformly split each video into multiple segments [13].

**Combining RNNs and CNNs.** There are several works that (like our own) combine the strengths of RNNs and CNNs. Most of these works relate to processing videos [12,18,40,49] and audio [19] using Recurrent Convolutional Networks (RCNs). RCNs typically operate on an input of image sequences (i.e., the frames of a video or spectrograms from audio recordings) where, at each timestep, a CNN extracts visual features from the image and feeds them to the RNN, which models the temporal dynamics of the sequence. RCNs have also been used to predict user states such as emotional valence and intentions from multi-channel EEG signals [28,47,50], where the input to the CNN encodes the spatial relationship among the EEG sensors placed on the user's head, along with their values. RCNs combine CNN and RNN at every timestep and therefore do not decouple the temporal from the spatial aspects of the data completely as done in VTNet. By providing a single scanpath to the CNN, VTNet processes the high-level spatial representation of the participant's overall activity related to a task, which complements more local temporal information about potential indicators of AD generated by the GRU from raw sequences.

An alternative way to combine a CNN and an LSTM is to use them in sequence instead of in parallel as VTNet does. Pascanu et al. [34] showed that LSTMs can learn better from high-level features of text embeddings extracted by CNNs, than from the raw embeddings themselves. This approach has been used mainly in NLP for sentiment analysis [2,11,29,44,46], and it does not seem to be useful for our goal of screening participants with AD because our ET data does not have such high dimensionality as word embeddings (each datapoint is a 6-dimensional vector), and reducing the dimensionality may result in a loss of possible patterns for the screening task.

## 3 Dataset

This paper relies on a dataset originally collected by Jang et al. [20] to build multimodal lightweight screening tools for AD. The data was collected from patients of a specialized memory clinic (either diagnosed with AD or showing early signs of mild cognitive impairments potentially leading to AD) and from control participants from the community (matched with the patient group based on sex and age).

During the study, participants were seated at a testing platform that had a Tobii-Pro X3-120 eye tracker (120 Hz sampling frequency) installed at the bottom of the screen to track gaze coordinates, head distance, and pupil data. The participants were given four tasks to complete: a pupil calibration task, a picture description task, a paragraph reading task, and a memory recall task that did not involve any visual elements. The study's full details are available in [20]. This paper utilizes ET data collected during the first three tasks:

- **Pupil Calibration:** Participants were asked to stare at a still target for 10-15 seconds (Figure 1A) to capture any square-wave jerks that are a hallmark of AD [33].
- **Picture Description:** Participants were asked to verbally describe the Cookie Theft picture (Figure 1B) from the Boston Diagnostic Aphasia Examination [16], a task that has been used extensively for assessing spontaneous speech in a variety of clinical settings [8], including AD [14,23,24].
- **Reading:** Participants were asked to read aloud a standardized paragraph (Figure 1C) from the International Reading Speed Texts (IReST), which is a collection of texts designed to assess reading impairments [42]. The 155-word paragraph described how plants and animals in hot and dry areas adapt to their environment in 9 sentences. The objective of this task was to capture common reading-task deficits associated with AD, such as reduced reading speed and increased word fixations or re-fixations.

Completing these three tasks took an average of 7 minutes. The final ET dataset used in this paper contains 75 control participants (avg. age = 62, std. dev. = 15) and 69 patients (avg. age = 72, std. dev. = 9). The sequence of raw ET samples for each user is represented by a 2D array (see Figure 2A), where the rows are the individual samples collected at 120Hz. Each sample is a 6-dimensional vector consisting of the gaze coordinates (Gx, Gy), the distance (HD) of the left and right eyes from the screen (used to estimate the head's distance from the screen), and the sizes of the left and right pupils (P). Table 1 shows the statistics for the dataset, including the number of datapoints for each task

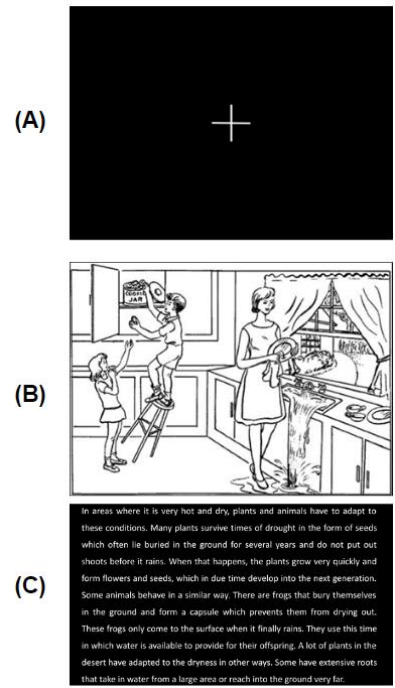

**Figure 1: Tasks used for collecting the AD dataset – (A) Pupil Calibration, (B) Picture Description, and (C) Reading.**

Table 1: Summary statistics of sequence lengths in the AD dataset

| Task | Group | N | Mean (Std. dev.) | Median | Min | Max |
|---|---|---|---|---|---|---|
| Pupil Calibration | Patient | 66 | 1461 (244) | 1405 | 1151 | 2317 |
|  | Control | 71 | 1369 (253) | 1362 | 106 | 2049 |
|  | Total | 137 | 1403 (249) | 1377 | 106 | 2317 |
| Picture Description | Patient | 67 | 7906 (4609) | 6488 | 1879 | 21861 |
|  | Control | 73 | 7974 (4659) | 7149 | 950 | 26103 |
|  | Total | 140 | 7948 (4626) | 6857 | 950 | 26103 |
| Reading | Patient | 67 | 8070 (3650) | 7265 | 375 | 20712 |
|  | Control | 73 | 6459 (1661) | 6476 | 1216 | 12066 |
|  | Total | 140 | 7080 (2719) | 6623 | 375 | 20712 |

and condition (patients and controls), as well as statistics on the length of the sequences[1].

## 4 Data preprocessing

The average sequence lengths for the three tasks are 1,403 for Pupil Calibration (std. dev.=249), 7,948 for Picture Description (std. dev.=4626), and 7,080 for Reading (std. dev.=2719), thus they are well above the length of 1,000 timesteps that is known to be suitable for LSTM-based models [27]. To address this issue, we adopted two data preprocessing steps. First, we cyclically split the ET sequences, as was done in [39]. The cyclical splitting process creates four separate datapoints from each original datapoint by assigning samples that are four steps apart to the same new datapoint in a cyclical manner (see Figure 2B). This process preserves the temporal structure of the ET data because there is little change between contiguous samples due to the high sampling rate while reducing the sequence length by a factor of 4. Additionally, the number of datapoints is increased by the same factor, as a form of data augmentation.

After cyclical splitting, the length of the sequences in the Pupil Calibration task is well below 1,000 (see the "Max" column for Pupil Calibration in Table 1, where the value should be divided by 4). However, this is not the case for the other two tasks. Hence, we experimented with applying a length cutoff to the sequences obtained from cyclical splitting to restrict their maximum length. We chose two cutoff values: the first cutoff value is 1,000, as it ensures that the maximum length of sequences never exceeds the threshold that is typically considered challenging for LSTM models [27]. The second cutoff value is 2000, to examine the effects of a less severe reduction in sequence length by including almost complete information from sequences with mean lengths. Figure 3 shows the distribution of sequence lengths after applying the 2,000-cutoff, which leads to 51 patients and 58 controls for the Picture Description task, and 57 patients and 70 controls for the Reading task having sequences above 1,000 timesteps.

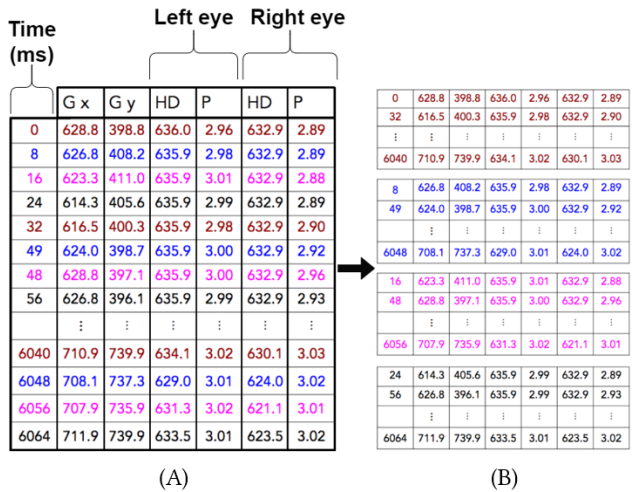

Figure 2: (A) An example of a datapoint, which is a sequence of ET samples (rows) from a given user. (B) The four distinct datapoints obtained from the datapoint in (A) through cyclical splitting.

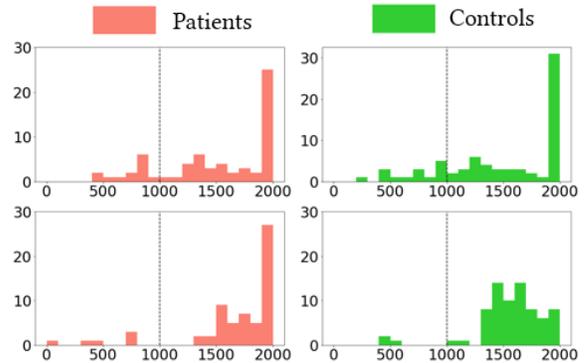

Figure 3: Distribution (Y axis) of sequence lengths (X axis) after applying the 2000 cutoff, for the Picture Description (top) and Reading task (bottom).

---

[1] These are obtained after removing outliers that are 3 std. dev. away from the mean from each task (7 for Pupil Calibration, 4 for Picture Description, and 4 for Reading).

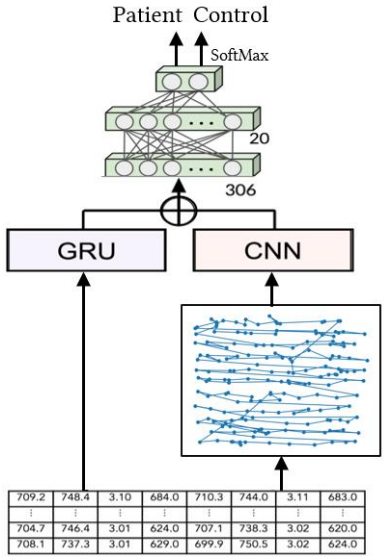

**Figure 4: The VTNet architecture.**

The application of the cutoff resulted in two variations of the Picture Description and Reading tasks' datasets, with each having a maximum sequence length of, respectively, 1,000 and 2,000 timesteps. We will use these two variations, as well as the dataset with no cutoff applied (all with cyclical splitting) to evaluate the performance of VTNet in distinguishing between patients and controls, as described in the next section.

## 5 VTNet architecture

The VTNet model architecture was first introduced in [39] and is presented in Figure 4. The model consists of a single-layer GRU sub-model and a two-layer CNN sub-model. The GRU and CNN sub-models operate independently. The GRU processes the sequences of raw ET samples, the CNN processes the corresponding spatial representation, namely the scanpath, that shows where fixations happened and the transitions between them (see as an example the input image to the CNN in Figure 4). The output of the GRU's 256-unit hidden state is concatenated with the 50-element vector output of the CNN to produce a single 306-sized vector. This vector is then passed to a simple neural network with one hidden layer and a SoftMax layer, which generates two outputs that indicate the model's confidence in classifying the input as either AD or control.

The VTNet hyperparameters used for this work are the same as in the original work [39], which discusses how this architecture was designed to be as simple as possible to deal with the limited size typical of ET datasets. The model is trained end-to-end as a single entity.

## 6 Evaluation of VTNet

### 6.1 Experimental Setup

Our evaluation aims to ascertain how VTNet compares to the best-performing non-DL classifiers from [20] in distinguishing between patients and controls. Therefore, following [20], we evaluate VTNet separately on each of the three tasks (Pupil Calibration, Picture Description, and Reading). For Pupil Calibration, VTNet is evaluated only on the full sequences since the length of the sequences here is less than 1,000 timesteps, as explained in Section 4. For the other two tasks, VTNet is evaluated on the full sequences, as well as on the sequences obtained by applying the length cutoffs of 1,000 and 2,000 timesteps. Hence, we label these three different VTNet models as *VTNet_full*, *VTNet_1000*, and *VTNet_2000* respectively.

For each task, the performance of the various VTNet models is compared against the best performing model among the non-DL classifiers tested in [20] (called baseline models from now on). It should be noted that the current AD dataset (described in Section 3) is larger than the version used [20] because participant recruitment is ongoing. Thus, we re-trained the non-DL models tested in [20] (Gaussian Naïve Bayes, Random Forest, Logistic Regression) on the current dataset, and selected the best performing model in each task (reported in Table 2) as a baseline for comparison with the VTNet models. All models are evaluated using 10 runs of 10-fold cross-validation (CV), and the results reported in the next section are the average of the 10 runs of 10-fold CV. Cross-validation is done across users, ensuring that no user contributes data points to both the training and test sets of a given fold. Cross-validation is also stratified so that the distribution of data points in each fold is kept similar to that of the dataset. For the non-DL models, we use the same hyperparameters as in [20] and we report the same performance metrics, which include:

1. **AUC** (Area Under Curve), which measures the accuracy of the classifier in distinguishing between patients and controls.

2. **Sensitivity** (or true positive rate), indicating the model's ability to detect patients.

3. **Specificity** (or true negative rate), indicating the model's ability to detect controls.

While AUC provides an overall performance measure, sensitivity and specificity are important in medical applications to estimate the likelihood of false negatives and false positives.

We formally compare model performances in each task by running a one-way MANOVA test with classifier type as the factor and the three performance metrics as dependent variables. Post-hoc comparisons are done with Tukey's HSD tests and statistical significance is reported for $p < 0.05$.

### 6.2 Results

Table 2 summarizes the performance of all tested models in each task. In Table 2, bold indicates the model with the highest numerical performance, whereas an asterisk indicates whether a specific VTNet model is statistically significantly better than the baseline (best performing) non-DL model. Table 3 summarizes the post-hoc comparisons where the differences for models with different underlines are statistically significant (e.g., Sensitivity of Baseline vs VTNet_full in the Pupil Calibration task), whereas differences for models with the same underline are not (e.g., AUC of Baseline vs VTNet_full in the Pupil Calibration task).

**Table 2: Performance of VTNet models trained on ET sequences with different maximum lengths and the corresponding best performing non-DL models, for each task.**

| Task | Classifier Type | AUC Mean (std. dev.) | Sensitivity Mean (std. dev.) | Specificity Mean (std. dev.) |
|---|---|---|---|---|
| Pupil Calibration | Gaussian Naïve Bayes | 0.71 (0.02) | **0.72 (0.02)** | 0.57 (0.04) |
|  | VTNet_full | 0.70 (0.01) | 0.64 (0.01) | **0.74 (0.02)** * |
| Picture Description | Random Forest | **0.75 (0.01)** | **0.65 (0.03)** | **0.72 (0.03)** |
|  | VTNet_1000 | 0.67 (0.01) | **0.65 (0.02)** | 0.66 (0.01) |
|  | VTNet_2000 | 0.63 (0.01) | 0.62 (0.02) | 0.63 (0.02) |
|  | VTNet_full | 0.58 (0.01) | 0.54 (0.02) | 0.66 (0.02) |
| Reading | Logistic Regression | 0.74 (0.03) | 0.59 (0.03) | 0.77 (0.01) |
|  | VTNet_1000 | 0.62 (0.01) | 0.63 (0.02) * | 0.63 (0.04) |
|  | VTNet_2000 | 0.73 (0.01) | 0.66 (0.01) * | 0.74 (0.01) |
|  | VTNet_full | **0.75 (0.01)** | **0.68 (0.01)** * | **0.78 (0.01)** |

For the Pupil Calibration task, the MANOVA shows a significant effect of the classifier on both sensitivity and specificity (for sensitivity: $F_{1,18}=136.774$, $p<.001$, partial $\eta2=.884$; for specificity: $F_{1,18}=192.823$, $p<.001$, partial $\eta2=.915$). Post-hoc comparisons (Table 3A) confirm that the baseline has higher sensitivity than VTNet, whereas VTNet has a higher specificity, with no difference in AUC scores.

For the Picture Description task, the MANOVA shows a significant effect of the classifier on all three performance metrics (for AUC: $F_{3,36}=676.844$, $p<0.001$, partial $\eta2=.983$; for sensitivity: $F_{3,36}=58.983$, $p<.001$, partial $\eta2=.831$; for specificity:

**Table 3: Statistical comparisons of models' performances with Tukey's HSD. Differences for models with the same underlines are not statistically significant, whereas differences for models with different underlines are.**

| A) Pupil Calibration | |
|---|---|
| AUC | Baseline > VTNet_full |
| Sensitivity | Baseline > VTNet_full |
| Specificity | **VTNet_full** > Baseline |

| B) Picture Description | |
|---|---|
| AUC | Baseline > VTNet_1000 > VTNet_2000 > VTNet_full |
| Sensitivity | Baseline > VTNet_1000 > VTNet_2000 > VTNet_full |
| Specificity | Baseline > VTNet_1000 > VTNet_full > VTNet_2000 |

| C) Reading | |
|---|---|
| AUC | VTNet_full > Baseline > VTNet_2000 > VTNet_1000 |
| Sensitivity | **VTNet_full** > VTNet_2000 > VTNet_1000 > Baseline |
| Specificity | VTNet_full > Baseline > VTNet_2000 > VTNet_1000 |

$F_{3,36}=25.096$, $p<.001$, partial $\eta2=.677$). Post-hoc comparisons (Table 3B) show that the baseline beats all VTNet models in terms of AUC and specificity. For sensitivity, the baseline and VTNet_1000 have equivalent performance and they outperform the other two VTNet models.

For the Reading task, the MANOVA shows a significant effect of the classifier on all three performance metrics (AUC: $F_{3,36}=125.352$, $p<0.001$, partial $\eta2=.913$; sensitivity: $F_{3,36}=51.244$, $p<.001$, partial $\eta2=.810$; specificity: $F_{3,36}=91.483$, $p<.001$, partial $\eta2=.884$). The post-hoc comparisons (Table 3C) show that all three VTNet models outperform the baseline in terms of sensitivity, with VTNet_full being the winning model. VTNet_full and the baseline are equivalent in specificity, and they outperform the other two VTNet models. For AUC, VTNet_full, VTNet_2000, and the baseline have equivalent performance and they outperform VTNet_1000.

### 6.3 Discussion

Based on overall performance (AUC) scores, no VTNet model outperforms the baseline models in any task. We hypothesized as a reason for this result that, despite the undertaken preprocessing steps, the sequences were still too long for the GRU sub-model to process. It is, however, interesting to observe the different trends between the Picture Description and the Reading tasks in terms of VTNet performance with different sequence lengths.

For the Picture Description task, the shorter the sequences the better, with VTNet_1000 outperforming VTNet_2000, which in turns outperforms VTNet_full, in both AUC and sensitivity. We observe the opposite trend in the Reading task. In terms of AUC, there is a non-significant trend of VTNet_full being better than VTNet_2000, which in turn is significantly better than VTNet_1000. There are similar but stronger trajectories for specificity and sensitivity, where the differences are statistically significant. These opposite trends suggest that in the Reading task, behaviors happen toward the end of the task that help distinguish between patients and control. For instance, it might be the case that as patients progress further in the paragraph, their reading impairments become increasingly evident, resulting in more discriminative ET behaviors that are captured partly with VTNet_2000 and fully with VTNet_full. In contrast, somehow the differences in how patients and controls visually process the Cookie Theft picture may get diluted as the task progresses, thus diminishing the ability of VTNet to discriminate between the two groups when looking at longer sequences. Further clarity on this point could be achieved by doing an offline analysis of gaze patterns at the end of sequences for both tasks. Given that none of our VTNet models outperform the baseline for AUC, we investigate if we can improve their performance by adding an attention layer, as discussed in the next section.

## 7 Adding an attention layer to VTNet

An attention layer computes the dot products of input sequences and learned weight vectors, producing attention scores that are normalized and used to weight the input sequences. As discussed in Section 2, there is evidence from both NLP and Computer Vision research that adding an attention layer to an LSTM-based model enables the model to focus on the most relevant parts of an input sequence, thus allowing it to capture long-term dependencies more effectively. Hence, in this section, we explore adding a self-attention layer before the GRU sub-model in VTNet. To implement this self-attention layer, we utilized PyTorch's (v1.13.0+cu117) default multi-head attention layer implementation. The dimension of this layer was set to 6 to match the dimensionality of our gaze data (see Figure 2A). The number of parallel attention heads was set to 1. This is because increasing the number of parallel attention heads increases the number of trainable parameters, which can result in overfitting when the dataset is small, as is the case in our work. Moreover, a smaller number of attention heads can also reduce the computational complexity of the model, resulting in more efficient training and evaluation.

### 7.1 Experimental Setup

To ascertain the effectiveness of augmenting the VTNet architecture with attention, we compare the augmented VTNet models against the original VTNet models and the non-DL baselines from [20]. As we did in Section 6, we perform this comparison for each of the three experimental tasks. For each task, we select the VTNet model that performed the best in the evaluation in Section 6, namely, VTNet_full for Pupil Calibration and Reading, and VTNet_1000 for Picture Description. These models augmented with attention are denoted with the suffix "_att" in the following sections. For example, *VTNet_2000_att* refers to the VTNet model with attention trained on the dataset with a maximum sequence length of 2,000 timesteps. The evaluation process is similar to that described in Section 6.1, where we utilize a one-way MANOVA test with classifier type as the factor and the three performance metrics as dependent variables to compare the relevant models. Similarly, post-hoc comparisons are performed using Tukey's HSD tests and statistical significance is reported for p < 0.05.

### 7.2 Results

Table 4 summarizes the results of this analysis. For the Pupil Calibration task, the MANOVA shows a significant effect on all three performance metrics (AUC: $F_{2,27}=137.134$, p<0.001, partial $\eta^2=.910$; sensitivity: $F_{2,27}=49.424$, p<.001, partial $\eta^2=.785$; specificity: $F_{2,27}=184.747$, p<.001, partial $\eta^2=.932$). Post-hoc comparisons (Table 5A), show a substantial improvement in performance with VTNet_full_att. This model now beats the baseline with an AUC of 0.78, which is a 9.8% increase (whereas VTNet_full is equivalent to the baseline), For sensitivity, VTNet_full_att is equivalent to the baseline (whereas VTNet_full is worse). For specificity, VTNet_full_att matches the performance of VTNet_full, which already beats the baseline. Furthermore, VTNet_full_att is a very balanced classifier, with 0.71 sensitivity, and 0.75 specificity (whereas VTNet_full has much better specificity than sensitivity with a difference of 10%).

For the Picture Description task, the MANOVA shows a significant effect on all three performance metrics (AUC: $F_{2,27}=192.702$, p<0.001, partial $\eta^2=.935$; sensitivity: $F_{2,27}=27.557$, p<.001, partial $\eta^2=.671$; specificity: $F_{2,27}=27.276$, p<.001, partial $\eta^2=.669$). Post-hoc comparisons (Table 5B) show that VTNet_1000_att outperforms the baseline for both AUC and sensitivity (while VTNet_1000 is either worse or equivalent), with sensitivity being especially impacted by reaching 0.7, which is a 7.7% increase from the baseline. For specificity, VTNet_1000_att is equivalent to the baseline where VTNet_1000 is worse. As was the case for the Pupil Calibration task, VTNet_1000_att is also balanced with 0.70 sensitivity and 0.73 specificity. In this task, VTNet_1000 is also balanced but with limited accuracies for both measures (0.65 and 0.66).

Table 4: Performance of the most promising VTNet model, its corresponding attention variant, and the baseline non-DL model, for each task.

| Task | Classifier Type | AUC Mean (std. dev.) | Sensitivity Mean (std. dev.) | Specificity Mean (std. dev.) |
|---|---|---|---|---|
| Pupil Calibration | Gaussian Naïve Bayes | 0.71 (0.02) | **0.72 (0.02)** | 0.57 (0.04) |
| | VTNet_full | 0.70 (0.01) | 0.64 (0.01) | 0.74 (0.02) * |
| | VTNet_full_att | **0.78 (0.01)** * | 0.71 (0.02) | **0.75 (0.01)** * |
| Picture Description | Random Forest | 0.75 (0.01) | 0.65 (0.03) | 0.72 (0.03) |
| | VTNet_1000 | 0.67 (0.01) | 0.65 (0.02) | 0.66 (0.01) |
| | VTNet_1000_att | **0.76 (0.01)** * | **0.70 (0.02)** * | **0.73 (0.02)** |
| Reading | Logistic Regression | 0.74 (0.03) | 0.59 (0.03) | 0.77 (0.01) |
| | VTNet_full | 0.75 (0.01) | 0.68 (0.01) * | 0.78 (0.01) |
| | VTNet_full_att | **0.78 (0.01)** * | **0.70 (0.01)** * | **0.80 (0.02)** * |

**Table 5: Statistical comparisons of models' performances with Tukey's HSD.**

| A) Pupil Calibration | |
|---|---|
| AUC | **VTNet_full_att** > Baseline > VTNet_full |
| Sensitivity | Baseline > VTNet_full_att > VTNet_full |
| Specificity | VTNet_full_att > VTNet_full > Baseline |

| B) Picture Description | |
|---|---|
| AUC | **VTNet_1000_att** > Baseline > VTNet_1000 |
| Sensitivity | **VTNet_1000_att** > Baseline > VTNet_1000 |
| Specificity | VTNet_1000_att > Baseline > VTNet_1000 |

| C) Reading | |
|---|---|
| AUC | **VTNet_full_att** > VTNet_full > Baseline |
| Sensitivity | VTNet_full_att > VTNet_full > Baseline |
| Specificity | **VTNet_full_att** > VTNet_full > Baseline |

For the Reading task, the MANOVA shows a significant effect on all three performance metrics (AUC: $F_{2,27}=10.484$, p<0.001, partial $\eta^2=.437$; sensitivity: $F_{2,27}=113.574$, p<.001, partial $\eta^2=.894$; specificity: $F_{2,27}=22.288$, p<.001, partial $\eta^2=.623$). The post-hoc comparisons (Table 5C) show that VTNet_full_att beats the baseline for all measures, and it is either better than (AUC and specificity) or equivalent (sensitivity) to VTNet_full. Interestingly, with 0.70 sensitivity and 0.80 specificity, this VTNet_full_att classifier is not as balanced as its counterparts in the Pupil Calibration and Picture Description tasks. This imbalance is due to a much higher specificity (0.80 compared to 0.75 in Pupil Calibration and 0.73 in Picture Description), whereas sensitivity is around 0.70-0.71 for all three models, showing that the attention layer mostly improves the model's ability to correctly classify control participants.

### 7.3 Discussion

Our results show that the addition of the attention layer enables the VTNet architecture to outperform the baseline models in all metrics for all tasks, except for sensitivity in the Pupil Calibration task where there is no statistical difference. These results indicate that the attention mechanism enables the GRU sub-model to better focus on critical parts of the ET sequences, despite their length.

One interesting question is whether adding the self-attention layer enhances VTNet's performance regardless of sequence length. To answer this question, we experimented with using VTNet augmented with attention on the confusion dataset from [39], where the ET sequences have a maximum length of 150 timesteps. We found that, with this confusion ET dataset, the VTNet models with and without attention had statistically equivalent performances on all metrics, suggesting that the addition of the attention layer to this architecture is not advantageous when the ET sequences are of manageable length. This is different than what is observed in NLP tasks, where attention helps even with sequences no longer than 120 tokens [6,38,48]. This difference could be due to a variety of reasons, including the nature of the classification task, type of data, and amount of information captured at any time step (e.g., a word arguably has higher information content than an individual raw gaze sample at a particular timestep), calling for further investigation on the relationship between all these factors, sequence lengths, and attention effectiveness.

## 8 Conclusions and Future Work

In this paper, we investigated whether VTNet, a model originally developed to classify user confusion from their eye-tracking (ET) data by processing in parallel a visual and temporal representation of the data, can also improve on the state-of-the-art results in classifying Alzheimer's Disease (AD). We addressed the challenge of long ET sequences by combining targeted length reduction with the addition of an attention layer to VTNet and showed that the results outperform the state-of-the-art for AD classification with ET data.

Our work has two contributions: first, the development of more accurate and lightweight classifiers for AD, and second, initial evidence of the generalizability of VTNet for leveraging ET data in different classification tasks.

Moving forward, we plan to experiment with building ensemble classifiers that combine VTNet for ET data and classifiers leveraging language data available in the AD dataset, as was done in Jang et al. [20] with non-DL models. We are especially interested in ascertaining if VTNet can be used to classify AD from speech data in the AD dataset by processing, in parallel, raw speech signals and their corresponding spectrograms, as is done for ET data. Additionally, we plan on testing VTNet on other ET datasets that have been used to predict user states such as learning [22] and affective valence [25].